\documentclass[10pt,twocolumn,letterpaper]{article}

\providecommand{\IfPDFManagementActiveF}[1]{}
\makeatletter
\@ifundefined{IfPDFManagementActiveF}{\def\IfPDFManagementActiveF#1{}}{}
\makeatother

\usepackage[pagenumbers]{cvpr} 

\usepackage{graphicx}
\usepackage{amsmath}
\usepackage{amssymb}
\usepackage{booktabs}
\usepackage[justification=centering]{caption}
\usepackage[ruled,vlined]{algorithm2e} 
\usepackage{algorithmicx}
\usepackage{float}
\usepackage{subcaption}
\usepackage{svg}
\usepackage{color}
\usepackage{enumitem}
\usepackage{tabularray}

\usepackage[pagebackref,breaklinks,colorlinks]{hyperref}

\usepackage[capitalize]{cleveref}
\crefname{section}{Sec.}{Secs.}
\Crefname{section}{Section}{Sections}
\Crefname{table}{Table}{Tables}
\crefname{table}{Tab.}{Tabs.}

\def\cvprPaperID{*****} 
\def\confName{CVPR}
\def\confYear{2026}

\begin{document}
	
	\title{Physics-Guided Fusion for Robust 3D Tracking of Fast Moving Small Objects}
	
	\author{
		Prithvi Raj Singh\thanks{Corresponding author} \textsuperscript{1}\and
		Raju Gottumukkala\textsuperscript{1} \and
		Anthony S. Maida\textsuperscript{1} \and
		Alan B. Barhorst\textsuperscript{1} \and
		Vijaya Gopu\textsuperscript{1, 2}\\
		\textsuperscript{1}UL Lafayette, \textsuperscript{2}Louisiana Transportation Research Center \\
		{\tt\small psingh8@mcneese.edu, V.Gopu@la.gov},
		{\tt\small \{raju, maida, barhorst\}@louisiana.edu}
	}
	
	\maketitle
	
	\begin{abstract}
		While computer vision has advanced considerably for general object detection and tracking, the specific problem of fast-moving tiny objects remains underexplored. This paper addresses the significant challenge of detecting and tracking rapidly moving small objects using an RGB-D camera. Our novel system combines deep learning-based detection with physics-based tracking to overcome the limitations of existing approaches. Our contributions include: (1) a comprehensive system design for object detection and tracking of fast-moving small objects in 3D space, (2) an innovative physics-based tracking algorithm that integrates kinematics motion equations to handle outliers and missed detections, and (3) an outlier detection and correction module that significantly improves tracking performance in challenging scenarios such as occlusions and rapid direction changes. We evaluated our proposed system on a custom racquetball dataset. Our evaluation shows our system surpassing kalman filter based trackers with up to 70\% less Average Displacement Error. Our system has significant applications for improving robot perception on autonomous platforms and demonstrates the effectiveness of combining physics-based models with deep learning approaches for real-time 3D detection and tracking of challenging small objects.
	\end{abstract}

	
	\section{Introduction}
	Advances in Computer Vision with state-of-the-art (SOTA) detection and tracking methods have led machines to better perceive the visual world leading to rapid growth in autonomous vehicles, augmented reality, and robotics among many fields. While state-of-the-art deep learning (DL)  models demonstrate impressive performance on general object detection tasks, they struggle significantly when confronted with fast-moving small objects (FMOs). This limitation represents a critical gap in current computer vision capabilities that restricts applications requiring precise tracking of small, dynamic targets. For reference, smaller objects can occupy an area between 1x1 pixels to 32x32 pixels as mentioned in ~\cite{kisantal2019augmentation}. In our research, the target small object occupies an area between 8x8 and 62x62 pixels in the frame. \par
	
	\begin{figure}[ht]
		\centering
		\includegraphics[width=\linewidth]{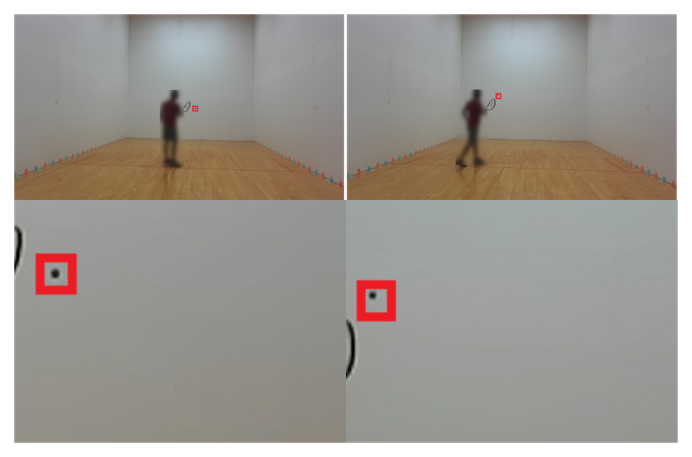}
		\caption{Illustration of our target object in an ideal racquetball game. Row 1 shows the ball as seen from a camera approximately 35 feet away from the front wall and Row 2 shows a zoomed-in picture of our target object. Our proposed approach improves tracking of the tiny ball.}
	\end{figure}
	
	\subsection{Problem Statement}
	Fast-moving small objects present unique challenges that conventional detection and tracking systems fail to address effectively:
	\begin{itemize}[itemsep=0.2em]
		\item \textbf{Limited Spatial Features}: Small objects (typically occupying between 1×1 and 32×32 pixels) have minimal discriminative features, making them difficult to distinguish from background or similar-sized elements.
		\item \textbf{Motion-Induced Blur}: Objects moving at high velocities create motion blur that further degrades already limited visual features.
		\item \textbf{Temporal Aliasing}: Standard camera frame rates often cannot adequately capture the rapid transitions of FMOs, leading to inconsistent appearance across consecutive frames.
		\item \textbf{Erratic Trajectory Changes}: Unlike larger objects with relatively predictable motion patterns, FMOs can undergo sudden directional changes, particularly after collisions with surfaces.
		\item \textbf{Occlusion Vulnerability}: The small visual footprint of these objects makes them especially susceptible to full occlusion, even by relatively small foreground elements.
	\end{itemize}
	
	Consider a racquetball with a true diameter of 57 millimeters (approximately 8-62 pixels in our captured frames depending on camera distance). These balls can travel at velocities exceeding 180 mph and change direction instantaneously upon collision with walls or the ground. Standard trackers like SORT~\cite{SORT}, BoTSORT~\cite{botsort}, and OCSORT~\cite{OCSORT} frequently experience tracking drift due to missed detections, particularly during these rapid directional changes. 
	
	\subsection{Limitations of Existing Approaches}
	Current approaches in small object detection and tracking like ~\cite{ju2020trajectory, golfball, tracknet, sun2020tracknetv2, yolo-z} primarily rely on a detect-then-associate paradigm. For small objects, this approach is fundamentally limited by:
	\begin{itemize}[itemsep=0.2em]
		\item \textbf{Detection Failures}: State-of-the-art detectors show significant performance degradation when applied to small objects, with precision and recall dropping by as much as 50\% compared to normal-sized objects.
		\item \textbf{Association Challenges}: Standard association algorithms struggle with the erratic motion patterns of FMOs, particularly after collisions, where linear motion assumptions break down.
		\item \textbf{Limited 3D Information}: Most existing datasets and methods focus solely on 2D tracking, omitting valuable depth information that could improve tracking robustness.
	\end{itemize}
	
	Hyperparameter optimization of modern detectors and trackers alone cannot overcome these fundamental limitations, necessitating a novel approach that specifically targets the unique characteristics of fast-moving small objects.
	
	\subsection{Our Contributions}
	This paper presents some key contributions to address these challenges:
	\begin{itemize}[itemsep=0.2em]
		\item \textbf{Comprehensive System Design}: We propose an end-to-end system for detection and tracking of fast-moving small objects that integrates RGB-D camera input with specialized detection and physics-based tracking modules.
		\item \textbf{Physics-Based 3D Tracking}: We introduce a novel tracking approach that incorporates kinematics motion equations to predict object trajectories even in challenging scenarios involving occlusions, erratic motion, and collisions.
		\item \textbf{Outlier Detection and Correction}: Our system includes a specialized module to identify and correct outliers and missed detections, significantly improving tracking robustness.
		\item \textbf{Depth Integration}: We develop techniques to effectively utilize depth information, enhancing tracking performance through 3D spatial awareness.
		\item \textbf{Specialized Dataset}: We contribute a new RGB-D dataset specifically focused on fast-moving small objects in a racquetball scenario, addressing the lack of suitable public datasets for this problem domain.
		\item \textbf{Comprehensive Evaluation}: We provide extensive comparative analysis between our physics-based approach and state-of-the-art trackers across multiple challenging scenarios.
	\end{itemize}
	
	Unlike previous work focused on applications like badminton~\cite{sun2020tracknetv2}, tennis~\cite{tracknet}, which track relatively larger objects with more predictable trajectories, our approach specifically targets smaller objects with higher velocities and more erratic motion patterns. Furthermore, we expand beyond the 2D tracking commonly used in these domains to incorporate full 3D tracking and trajectory forecasting.
	
	The remainder of this paper is organized as follows: Section 2 reviews related work in object detection, tracking, and small object detection. Section 3 details our proposed system design. Section 4 elaborates on our physics-based 3D tracking and forecasting approach. Section 5 presents experimental results and analysis. Finally, Section 6 concludes the paper and discusses directions for future work.
	
	\section{Related Works}
	The field of object detection and tracking has seen significant advances with the emergence of deep learning. However, several specific challenges remain, particularly in the domain of small, fast-moving object detection and tracking. This section reviews the relevant works in three key areas: small object detection, tracking methods for dynamic objects, and specialized approaches for sports applications.
	
	\subsection{Small Object Detection}
	Small Object Detection (SOD) presents unique challenges compared to general object detection due to the limited discriminative features available in small objects. Despite the progress in general object detection with architectures like YOLOv5 ~\cite{khanam2024yolov5deeplookinternal}, YOLOv7 ~\cite{wang2023yolov7}, YOLOv8 ~\cite{yaseen2024yolov8indepthexplorationinternal}, Fast R-CNN ~\cite{fastrcnn} , and Faster R-CNN ~\cite{fasterrcnn}, these methods show significant performance degradation when applied to small objects. 
	
	Benjumea et al. propose YOLO-Z~\cite{yolo-z}, which improves YOLOv5's small object detection capabilities by modifying the architecture to better preserve small object features through the detection pipeline. Their approach demonstrates improved performance but remains limited to relatively slow and non moving objects in autonomous driving scenarios. 
	
	A comprehensive survey by Cheng et al.~\cite{surveySOD} highlighted the limitations of existing SOD datasets and benchmarks, noting that most focus on stationary or slowly moving objects in aerial or driving scenarios. Their analysis revealed that even state-of-the-art detectors experience a 30-50\% performance drop when applied to objects smaller than 32×32 pixels. 
	
	Kisantal~\cite{kisantal2019augmentation} explored data augmentation techniques specifically for small objects, showing that strategic oversampling and magnification of small objects during training can improve detection performance. However, their approach did not address the additional challenges introduced by rapid erratic motion. 
	
	The datasets presented in works by Ding et al.~\cite{DOTA}, Yu et al.~\cite{tinyperson}, and Zhu et al.~\cite{visDrone} focus primarily on stationary small objects or scenes where camera movement creates relative motion. These datasets do not capture the unique challenges presented by objects with intrinsic high-speed motion and frequent direction changes, which is the focus of our work. 
	
	\subsection{Object Tracking Methods}
	Object tracking approaches can be categorized into two groups: traditional feature-based methods and modern tracking-by-detection frameworks.
	
	Traditional methods like SIFT-based trackers~\cite{SIFT1, sift2,sift3,sift4} and mean-shift trackers ~\cite{meanshift1,meanshift2,meanshift3} rely on distinctive visual features or color distributions. Fazli et al.~\cite{sift4} combined SIFT features with color information for more robust tracking, while Comaniciu et al.~\cite{meanshift1} and Vojir et al.~\cite{meanshift2} proposed adaptive mean-shift approaches to handle appearance changes. However, these methods struggle significantly with small objects due to the limited features available and are particularly vulnerable to the motion blur that characterizes fast-moving objects. 
	
	Modern tracking-by-detection frameworks have become dominant in recent years, with various implementations of Kalman filter-based approaches showing strong performance on benchmark datasets. Simple Online and Real-time Tracking (SORT) ~\cite{SORT} pioneered the efficient combination of Kalman filtering for motion prediction with the Hungarian algorithm~\cite{kuhn1955hungarian} for detection association. Subsequent improvements include DeepSORT ~\cite{deepSORT}, which incorporates appearance features, and more recent variants like OCSORT ~\cite{OCSORT}, which enhances association reliability through observation-centric design. 
	
	ByteTrack ~\cite{bytetrack} addressed missed detections by incorporating low-confidence detections into the tracking process, while StrongSORT ~\cite{strongsort} combined the strengths of multiple approaches into an integrated framework. BoTSORT ~\cite{botsort} further improved association accuracy through better integration of appearance and motion cues. 
	
	Despite these advances, these trackers fundamentally assume relatively linear motion patterns between frames and struggle with the erratic trajectories characteristic of fast-moving objects after collisions. Additionally, they operate primarily in the 2D image space without leveraging depth information, which can be crucial for accurate 3D trajectory modeling. 
	
	More specialized approaches like COMET ~\cite{marvasti2020comet} have shown improved performance for small objects by incorporating context-aware modules and attention mechanisms to enhance discriminative features. However, these methods still face challenges with very fast motion that causes significant appearance changes between consecutive frames.
	
	\subsection{Approaches for Sports Application}
	Sports applications have driven specific research into fast-moving object tracking, though most existing work focuses on objects larger and slower than our target application. 
	
	TrackNet~\cite{tracknet} proposed a specialized architecture for tracking shuttlecocks in badminton videos. Their approach combines convolutional and deconvolutional layers with Gaussian heatmap generation to precisely localize the shuttlecock across frames. While effective for detection, their system operates at reduced frame rates that limit real-time application. 
	
	Sun et al.~\cite{sun2020tracknetv2} extended this approach with TrackNetV2, improving shuttlecock detection through architectural enhancements. However, both approaches operate solely in 2D space without incorporating depth information and target objects that are significantly larger and slower than racquetballs. 
	
	Zhang et al.~\cite{golfball} presented a system for golf ball tracking that combines Kalman filtering with cropped region processing to improve detection efficiency. Their approach demonstrates some similarities to our work in using motion models to guide detection, but their focus on golf balls presents fewer challenges for tracking. Golf follow more predictable trajectory and is unlikely to have erratic trajectory changes.
	
	These sports-focused approaches all operate in 2D space only, lacking the depth dimension that can significantly enhance tracking robustness. Additionally, they primarily target objects with more predictable trajectories and lower speeds compared to racquetballs, which can exceed 180 mph and change direction instantaneously upon collision.
	
	\subsection{Gap Analysis and our Approach}
	Based on our research~\cite{psingh2025} of existing works, we identify several key gaps that our work addresses:
	\begin{itemize}[itemsep=0.2em]
		\item \textbf{3D vs. 2D Tracking}: Existing approaches predominantly operate in 2D image space, while our work incorporates depth information from RGB-D cameras to enable full 3D tracking and trajectory modeling.
		\item \textbf{Object Speed and Size}: Most existing research targets either small stationary objects or larger moving objects. Our work specifically addresses very small (8×8 to 62×62 pixels) objects moving at extreme speeds ($\geq$ 180 mph).
		\item \textbf{Physics-Informed Tracking}: While Kalman filters provide statistical motion estimation, they struggle with the non-linear motion patterns resulting from collisions. Our physics-based approach explicitly models these interactions for more accurate trajectory prediction.
		\item \textbf{Erratic Motion Handling}: Existing trackers assume relatively smooth motion between frames, while our approach specifically addresses the erratic direction changes characteristic of objects like racquetballs after collision events.
	\end{itemize}
	
	By addressing these gaps, our work advances the state of the art in fast-moving small object detection and tracking, particularly for applications requiring precise 3D trajectory modeling in complex environments with frequent occlusions and collisions.
	
	\section{Proposed System Design}
	Fast-moving small objects present unique detection and tracking challenges that require a specialized system architecture. We propose an integrated approach that combines deep learning-based detection with physics-based tracking and depth information to achieve robust 3D tracking performance. Table ~\ref{OverallTracking} presents our system's performance.
	
	\subsection{Overview of System Architecture}
\begin{figure*}[ht]
		\centering
		\includesvg[width=\textwidth]{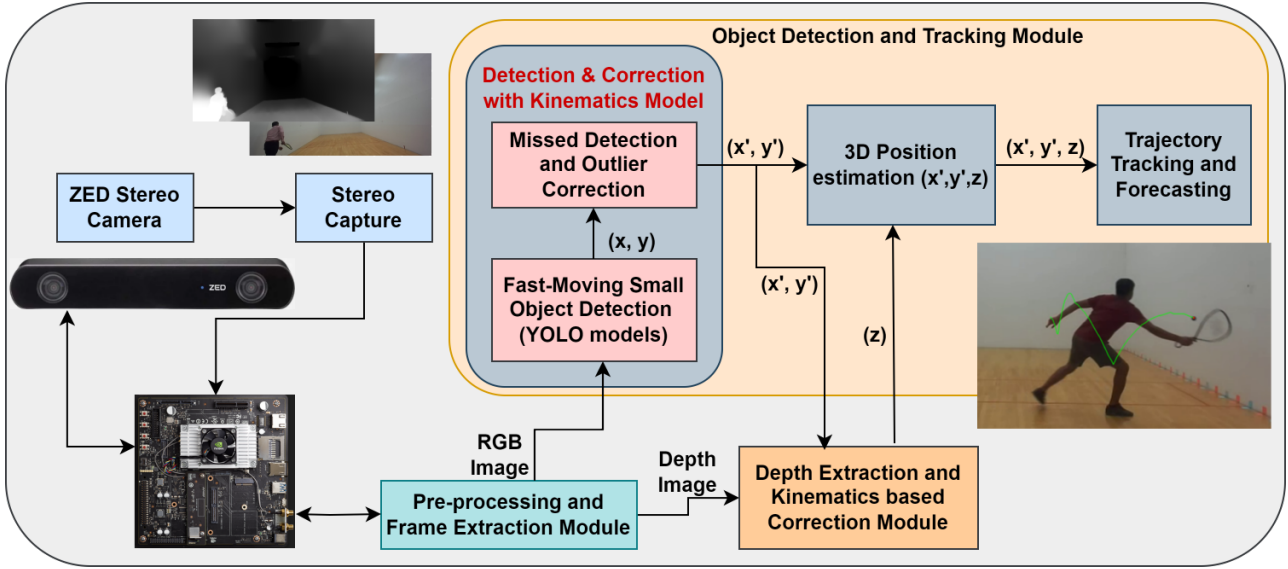}
		\caption{System Design for ball detection, tracking, and forecasting. The design shows the end-to-end process of our 3D perception system. The highlighted parts of the design represent different sub-processes or modules. We are using an NVIDIA Jetson TX2 as our main computing device for both collecting the dataset and extracting the 3D position of the ball.}
		\label{fig:systemdesign}
	\end{figure*}

	Our proposed system (illustrated in figure~\ref{fig:systemdesign}) consists of four primary modules, pre-processing and frame extraction module, object detection module, depth extraction and correction module, 3D object tracking and forecasting module, that operate sequentially to form a complete perception system. Each of these modules are presented in a detail in subsequent section. These modules work in conjunction to overcome the limitations of traditional approaches when applied to fast-moving small objects.
	
	\subsection{Pre-processing and Frame Extraction}
	The pre-processing module serves as the system's data ingestion component, handling the stereo video input from the RGB-D camera. The module serves key functions like frame synchronization, frame extraction, and resolution standardization. The frame synchronization ensures temporal alignment between RGB and Depth frames. Frame extraction seperates RGB frames from the associated depth maps and resolution standarization resizes inputs to standarized dimensions for consistent processing. 
	
	The module can extract either left or right RGB frames from stereo video based on calibration parameters. The synchronized depth maps contain raw depth measurements for each pixel, which are later processed to extract precise object distance.
	
	\subsection{Object Detection Module}
	The object detection module is localizes object (racquetball) within each RGB frame. After evaluating several state-of-the-art detection approaches (detailed in Section 5.3), we chose YOLOv8 as our primary detector due to its optimal balance between accuracy and inference speed. We enhanced the detector's performance for small objects using techniques like model fine-tuning, anchor optimization, data augmentation strategy, loss function weighting. 
	
	The detection module outputs bounding box coordinates (x, y, width, height) for each detected ball, along with a confidence score. The center point coordinates ($x'$, $y'$) are calculated and then passed to the depth extraction module.
	
	\subsection{Depth Extraction and Correction Module}
	The depth extraction module interfaces with the RGB-D camera's depth stream to retrieve the Z-coordinate (depth) corresponding to the detected object's center point. Here we address several challenges specific to depth sensing for small, fast-moving objects like depth sampling, noise filtering, missing value handling and ensuring temporal consistency. For depth sampling, we extract the depth value at the center coordinates ($x$, $y$) of the detected bounding box. We address cases where depth information is unavailable (NaN or infinite values) due to sensor limitations, occlusions, or surface properties. Temporal consistency ensures depth measurements are consistent across consecutive frames. The approx depth resolution is 1920x1080. 
	
	When valid depth information is unavailable, the module employs our physics-based model (detailed in Section 4.2) to estimate the missing Z-coordinate based on the object's previous trajectory and kinematic constraints. This allows the system to maintain tracking continuity even during temporary depth sensing failures. The output of this module is a 3D position (x, y, z) of the detected object, which forms the input for the tracking and forecasting module.
	
	\subsection{System Integration and Data Flow}
	The system processes data through interconnected stages. First, the RGB-D camera captures synchronized color and depth frames. Then, the pre-processing module extracts and prepares these individual frames for further analysis. The object detection module identifies the ball's location within the RGB frame, while the depth extraction module retrieves its corresponding Z-coordinate. Using this data, the 3D tracking and forecasting module associates detections across frames and predicts future positions of the ball. Finally, the results are fed back into the system to refine and improve detection and tracking accuracy in subsequent cycles.
	
	This integrated approach enables robust tracking performance even in challenging scenarios involving occlusions, rapid direction changes, and temporary detection failures. The system maintains a frame-by-frame record of the object's 3D position, which can be used for trajectory analysis, behavior prediction, or integration with robotic control systems.
	
	\section{Physics-based 3D Tracking and Forecasting}
	Any object’s movement can be modeled using the general laws of physics. Traditional tracking approaches like Kalman filter-based methods assume relatively constant velocity or acceleration between frames, which proves inadequate for objects exhibiting rapid directional changes after collisions. Our physics-based tracking approach explicitly models the kinematics and collision dynamics of fast-moving objects to achieve superior tracking performance and trajectory prediction.
	
	\label{sec:3Dtracking}
	\begin{figure}[ht]
		\centering
		\includesvg[width=\linewidth]{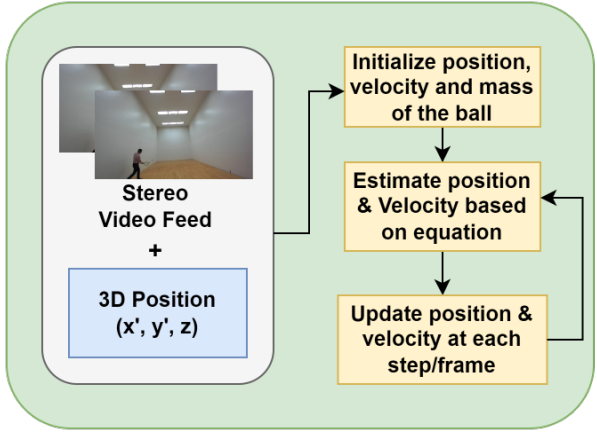}
		\caption{Outline of the physics-based tracker.}
		\label{GeneralOutline}
	\end{figure}
	
	\subsection{Tracking Methodology}
	Figure~\ref{GeneralOutline} illustrates the architecture of our physics-based tracker, which integrates detection outputs with a kinematics motion model. The tracker incorporates several interconnected components like state estimation, trajectory forecasting, detection association, collision detection and response, outlier rejection and correction. The state estimation module is responsible for maintaining the object's position, velocity, and acceleration at any given time. Using kinematics motion equations, the trajectory forecasting module predicts future object positions. The detection association component ensures new detections are accurately matched to these predicted positions. To account for dynamic scenarios, the collision detection and response module identifies collision events and models the resulting directional changes. The outlier rejection and correction module effectively identifies and addresses errors in detection to enhance the overall tracking accuracy. 
	
	Unlike purely learning-based approaches, our physics-based tracker incorporates knowledge about object motion, enabling more accurate predictions during challenging scenarios such as occlusions or detection failures.
	
	\subsection{Kinematics Motion Model}
	While a complete dynamics model would account for all forces acting on the object, including air resistance, friction, and spin effects, such models require extensive parameter tuning and computational resources. Our proposed kinematics-based approach provides an effective balance between accuracy and computational efficiency by focusing on the fundamental motion equations while incorporating the critical aspects of collision response. We formulate the kinematics motion model based on the explanation in ~\cite{everett2004dynamics}. 
	
	For an object in free flight (between collisions), the position at time $t$ can be determined using standard kinematics equations below:
	\begin{equation}
		z(t) = z_{0}+V_{z_{0}}t - 1/2gt^{2} 
		\label{zvalue}
	\end{equation}
	\begin{equation}
		y(t) =  y_{0}+ V_{y_{0}}t - 1/2gt^{2}
		\label{yvalue}
	\end{equation}
	\begin{equation} 
		x(t) = x_{0}+ V_{x_{0}}t
		\label{xvalue}
	\end{equation}
	
	Where $(x,y,z)$ represents the position at time $t$, $(x_0, y_0, z_0)$ are the initial positions, and $(V_{x0}, V_{y0}, V_{z0})$ are the initial velocities in each direction. The gravitational acceleration $g$ affects only the $y$ and $z$ components, depending on camera orientation.
	
	The velocity between consecutive positions is the ratio of displacement over time as shown in 4:
	\begin{equation}
		velocity _{i \rightarrow j} = \frac{displacement_{i\rightarrow j}}{t_{i\rightarrow j}}
		\label{Velocity1}
	\end{equation}
	
	\subsection{Collision Response Modeling}
	A key advantage of our physics-based approach is the explicit modeling of collision events, which represent critical points where traditional tracking methods often fail. When the ball collides with a surface (wall, floor, ceiling), its trajectory changes abruptly in a manner that depends on the collision dynamics.
	
	We model these collisions using the coefficient of restitution (COR or $e_r$), which characterizes the elasticity of the collision. The COR is defined as the ratio of the relative velocity after collision to the relative velocity before collision, $e_r = \frac{|\vec{V}_{a}|}{|\vec{V}_{b}|}$. Here $\vec{V_{a}}$ is velocity after impact and $\vec{V_{b}}$ represents velocity before or the previous known velocity. In 3D space, velocity before and after can be modeled using equations ~\ref{Vel_after} and ~\ref{Vel_before}.
	
	\begin{equation}
		\vec{V_{a}} = \vec{V}_{a_{1}\hat{x}} + \vec{V}_{a_{2}\hat{y}} + \vec{V}_{a_{3}\hat{z}}
		\label{Vel_after}
	\end{equation}
	
	\begin{equation}
		\vec{V_{b}} = \vec{V}_{b_{1}\hat{x}} + \vec{V}_{b_{2}\hat{y}} + \vec{V}_{b_{3}\hat{z}} 
		\label{Vel_before}
	\end{equation}
	
	For a perfectly elastic collision, $e_r = 1$, indicating complete conservation of kinetic energy. Real-world objects like racquetballs have $e_r$ values slightly below 1 (typically 0.85-0.95), reflecting minor energy losses during collision. When a collision is detected (based on position relative to known boundaries or sudden direction changes), the post-collision velocity is calculated as: $\vec{V}_{a} = e_{r} * \vec{V}_{b}$. Because the racquetball is elastic, total momentum is conserved.
	
	With the direction component adjusted based on the collision surface normal. For a flat surface collision, the velocity component perpendicular to the surface reverses direction while maintaining its magnitude (scaled by $e_r$), while components parallel to the surface remain unchanged. 
	
	Based on empirical testing with racquetball dynamics, we set $e_r = 0.95$ in our implementation, which accurately models the high elasticity of racquetballs while accounting for very small energy loss.
	
	\subsection{Tracking Algorithm Implementation}
	Algorithm~\ref{alg:ball-tracking} presents our complete tracking approach, which integrates the kinematics motion model with detection inputs to achieve robust tracking even during detection failures or outliers. The algorithm: (a) initializes tracking with initial detected position, (b) predict next position using the kinematics model, (c) associate new detections with predictions based on spatial proximity, (d) reject outliers if the distance between detection and prediction exceeds a threshold, (e) replace missing detections with physics-based predictions, (f) update velocity and position states after incorporating measurements, and (g) apply collision response when boundary interactions are detected. This approach shows several advantages over traditional approaches:
	\begin{itemize}[itemsep=0.2em]
		\item Outlier Rejection: By comparing detections against physics-based predictions, the algorithm can identify and reject false detections that would otherwise disrupt tracking.
		\item Occlusion Handling: During occlusions or missed detections, the physics model maintains tracking continuity with predicted positions.
		\item Collision Awareness: Explicit modeling of collision events enables accurate trajectory prediction even during the rapid direction changes that typically cause tracking failures.
	\end{itemize}
	
	\begingroup
	\small
	\setlength\emergencystretch{1.5em}
	\raggedright
	\sloppy
	\begin{algorithm}[htp]
		\caption{Ball Detection \& Tracking Algorithm}
		\label{alg:ball-tracking}
		\textbf{Data:} 
		\begin{itemize}[itemsep=0.2em]
			\item Stereo Video
			\item Initial Parameters
		\end{itemize}
		
		\textbf{Input variables:}
		\begin{itemize}[itemsep=0.2em]
			\item Initial\_pos: initial position $(x, y)$ of the ball.
			\item yolo\_pos: position $(x, y)$ of the ball as detected using YOLO.
			\item pred\_pos: predicted position of the ball using kinematics model 
			\item Threshold: a fixed number representing the maximum allowed distance between yolo\_pos and pred\_pos.
			\item $e_{r}$: coefficient of restitution.
		\end{itemize}
		
		\textbf{Result:} Predicted and Tracked Positions
		
		\While{frames in video}{
			\textbf{Perform Object Detection} to find the racquetball\;
			\textbf{Find center} $(x, y)$ of the detected bounding box\;
			Initial\_pos = yolo\_pos\;
			Predict the next position (x, y);
			
			\textbf{Continuously perform object detection} and assign $(x, y)$ to yolo\_pos\;
			
			\If{distance(yolo\_pos, pred\_pos) $>$ Threshold}{
				yolo\_pos = pred\_pos\;
			}
			
			\If{yolo\_pos is NaN }{
				yolo\_pos = pred\_pos\;
			}
			
			\If{ball strikes wall or ground}{
				vel = $e_{r}$ * vel\;}
			Update position and velocity for the next frame\;
			Construct Trajectory\;
		}
	\end{algorithm}
	\fussy
	\endgroup
	
	\subsection{Depth Extraction and Correction for 3D Tracking}
	Incorporating the depth dimension is critical for accurate 3D trajectory modeling but presents unique challenges. Algorithm~\ref{alg:depth-correction} details our approach to depth extraction and correction. 
	
	\begingroup
	\small
	\setlength\emergencystretch{1.5em}
	\raggedright
	\sloppy
	\begin{algorithm}[htp]
	\caption{Depth Extraction \& Correction}
	\label{alg:depth-correction}
		\textbf{Input:}
		\begin{itemize}[itemsep=0.2em]
			\item Center $(x, y)$, of the detected bounding box.
			\item Respective Depth Image
		\end{itemize}
		\textbf{Output:} 3D position (x, y, z) of the ball. \\
		\textbf{Procedure:} \\
		\While{frames in a video}{
			\textbf{For $(x, y)$} of the bounding box \\
			-- call ZED API for depth $z$ \\
			
			\If {$z$ is nan or inf}{ $z(t) = z_{0} + V_{z_{0}}t - \tfrac{1}{2}g t^{2}$ }
			\begin{itemize}
				\item Get the 3D position $(x, y, z)$ of the ball
				\item Enable 3D tracking and forecasting
		\end{itemize}}
	\end{algorithm}
	\fussy
	\endgroup

	The depth extraction process works as follows:
	\begin{itemize}[itemsep=0.2em]
		\item For each detected object center (x, y), query the corresponding depth map
		\item Check for invalid depth values (NaN or infinite values)
		\item If valid depth is available, incorporate it into the 3D position
		\item If depth is invalid or unavailable, estimate it using the kinematic model: $z(t) = z_{0}+V_{z_{0}}t - 1/2gt^{2}$
	\end{itemize}
	
	This approach ensures continuous 3D tracking even when depth measurements are temporarily unavailable due to sensor limitations, occlusions, or surface properties that challenge depth sensing (such as highly reflective or transparent surfaces). The integration of depth information significantly enhances tracking performance by enabling true 3D trajectory modeling, providing an additional dimension for outlier detection and validation, allowing more accurate collision prediction with 3D boundaries, and supporting applications requiring precise spatial awareness. \newline
	
	Together, these algorithms form a robust tracking system capable of maintaining accurate 3D trajectories for fast-moving small objects even in challenging scenarios involving occlusions, rapid direction changes, and detection uncertainties.
	
	\section{Results and Analysis}
	We conducted comprehensive experiments to evaluate the performance of various object detection and tracking methods on our specialized fast-moving small object dataset. This section presents our experimental setup, evaluation metrics, and detailed performance analysis. 
	
	\subsection{Dataset Collection and Preparation}
	To address the lack of suitable public datasets for fast-moving small object tracking, we collected and annotated a specialized dataset using a ZED 2i stereo RGB-D camera. Our dataset consists of 12 video sequences, each lasting 5 minutes, capturing racquetball gameplay. It offers resolutions of 720p and 1280p with synchronized RGB and depth frames, totaling over 80,000 frames. Ten thousand frames were manually annotated for training and validation purposes. 
	
\begin{figure*}[htp]
		\centering
		\includegraphics[width=0.9\textwidth]{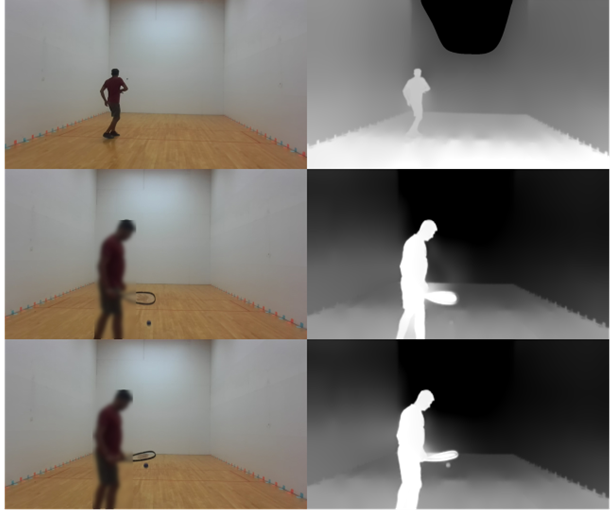}
		\caption{Sample of our RGB-D dataset. The figure shows the RGB image and its corresponding depth image. We used the RGB frame and its matching depth image to design our 3D tracking and trajectory forecasting system. }
		\label{datasetSample}
	\end{figure*}

	The dataset captures racquetballs traveling at speeds up to 180 mph with frequent direction changes due to collisions with walls, floor, and ceiling. Figure~\ref{datasetSample} shows sample frames from our dataset, illustrating both RGB and corresponding depth information. The racquetball diameter is 57mm, appearing between 8×8 and 62×62 pixels depending on distance from the camera. This small size, combined with high speed, makes detection particularly challenging compared to existing sports datasets. 
	
	Data annotation was performed using the \textit{makesense.ai} online tool, with bounding boxes drawn around the racquetball in each frame. We split the annotated frames into 80\% training and 20\% validation sets, ensuring that consecutive frames were split between sets to maintain temporal consistency. We tested the trained model on remaining frames from our overall dataset. We plan on releasing the dataset in future once we have the Institutional Review Board (IRB) approval. 
	
	\subsection{Evaluation Metrics}
	\subsubsection{Object Detection Metrics}
	To evaluate object detection performance, we used standard metrics like precision, recall, mean Average Precision (mAP$@$0.5), mAP$@$0.5:0.95, F1-score. Detailed description of these metrics can be found in ~\cite{Metrics}
	\begin{itemize}[itemsep=0.1em]
		\item \textbf{Precision}: The ratio of correct detections to total detections
		\item \textbf{Recall}: The ratio of correct detections to the total number of ground truth objects
		\item \textbf{mean Average Precision (mAP@0.5)}: AP calculated at IoU threshold of 0.5
		\item \textbf{mAP@0.5:0.95}: Average AP over multiple IoU thresholds from 0.5 to 0.95
		\item \textbf{F1 Score}: Harmonic mean of precision and recall
	\end{itemize}
	
	\subsubsection{Tracking Metrics}
	To evaluate object tracking performance, we used two primary metrics - ADE and AMD. These metrics provide complementary assessment of tracking accuracy, with ADE measuring absolute spatial error and AMD accounting for prediction uncertainty. Lower ADE and AMD scores are better.
	\begin{itemize}[itemsep=0.2em]
		\item \textbf{Average Displacement Error (ADE)}: Mean Euclidean distance between predicted trajectories and ground truth positions, calculated as:
		\[{ADE} = \frac{\sum_{1}^{n}{\sqrt{(x_{2}- x_{1})^2 + (y_{2}- y_{1})^2 + (z_{2}- z_{1})^2}}}{N_{traj}}\]
		\item \textbf{Average Mahalanobis Distance (AMD)}: AMD is a mean of Mahalanobis Distance (MD) between all point pairs. Mahalanobis Distance is the distance between pair points in multivariate space. MD considers correlations between variables, as well as the variances of the variable. For two points x and y in a p-dimensional space, MD is, 
		\[MD_{i}(\hat{\mu},\hat{\sum },p)=\sqrt{(p-\hat{\mu})^{T}\hat{\sum }^{-1}(p-\hat{\mu})}\]
		For AMD, we sum all MD values and divides by the total number of point pairs N.
		\[AMD = \frac{1}{N}\sum_{i=1}^{N}MD_{i} \]
	\end{itemize}
	
	\subsection{Object Detection Performance}
	We evaluated several state-of-the-art object detection methods after fine-tuning them on our dataset. Table~\ref{PerformAllDet} presents the detection performance metrics for each model. 
	
	YOLOv5 achieved the highest precision (0.856) and F1 score (0.85), while YOLOv8 demonstrated strong balanced performance across metrics with 0.84 precision and 0.80 recall. TrackNetV2, despite being specialized for small sports objects, achieved good precision (0.85) but lower recall (0.72), indicating missed detections of the fast-moving racquetball. 
	
	Faster R-CNN, while achieving respectable precision (0.826), showed significantly lower mAP scores, suggesting challenges with consistent detection across varying object sizes and positions.
	
	
	\begin{table}[!ht]
		\centering
		\resizebox{\columnwidth}{!}{%
			\begin{tabular}{lccccc}
				\toprule
				\textbf{Method} & \textbf{Precision} & \textbf{Recall} & \textbf{mAP@0.5} & \textbf{mAP@0.5:0.95} & \textbf{F1} \\
				\midrule
				\textbf{YOLOv5}      & \textbf{0.856} & \textbf{0.845} & \textbf{0.790} & \textbf{0.313} & \textbf{0.850} \\
				YOLOv7      & 0.700 & 0.630 & 0.640 & 0.225 & 0.660 \\
				YOLOv8      & 0.840 & 0.800 & 0.760 & 0.300 & 0.820 \\
				Faster-RCNN & 0.826 & 0.780 & 0.650 & 0.230 & 0.800 \\
				TrackNetv2  & 0.850 & 0.720 & 0.680 & 0.256 & 0.770 \\
				\bottomrule
			\end{tabular}
		}
		\caption{Performance of different object detection methods. Best values are highlighted in bold.}
		\label{PerformAllDet}
	\end{table}
	
	\subsubsection{Inference Speed Analysis}
	Table ~\ref{InfSpeed} compares the inference speed of different object detectors at two frame sizes: 640×384 (standard YOLO input size) and 1248×704 (close to our original capture resolution). This analysis is critical for real-time applications. 
	
	YOLOv8 demonstrated superior performance with 69 FPS at 640×384 and 41 FPS at 1248×704, representing a 2-4× speed advantage over most competitors. Faster R-CNN, despite reasonable detection accuracy, proved impractical for real-time applications with only 3-5 FPS.
	Based on both accuracy and speed considerations, YOLOv8 offers the optimal balance for our application, providing sufficient accuracy while maintaining real-time performance even at higher resolutions.
	
	
	\begin{table}[!ht]
		\centering
		\resizebox{\linewidth}{!}{%
			\begin{tabular}{lcccc}
				\toprule
				& \multicolumn{2}{c}{\textbf{Image Size (640×384)}} & \multicolumn{2}{c}{\textbf{Image Size (1248×704)}} \\
				\cmidrule(lr){2-3} \cmidrule(lr){4-5}
				\textbf{Method} & Latency (ms) & FPS (Approx.) & Latency (ms) & FPS (Approx.) \\
				\midrule
				YOLOv5      & 25.0 & 40 & 58.7 & 17 \\
				YOLOv7      & 28.6 & 34 & 59.0 & 17 \\
				\textbf{YOLOv8}      & \textbf{14.5} & \textbf{69} & \textbf{24.5} & \textbf{41} \\
				Faster-RCNN & 200.0 & 5 & 320.0 & 3 \\
				TrackNetv2  & 48.0 & 20 & 52.4 & 16 \\
				\bottomrule
			\end{tabular}
		}
		\caption{Inferencing speed of object detectors at given frame sizes (width × height). Best values are highlighted in bold.}
		\label{InfSpeed}
	\end{table}
	
	\subsection{Tracking Performance Analysis}
	We compared our physics-based tracker against five state-of-the-art tracking methods: DeepOCSORT~\cite{DeepOC}, OCSORT~\cite{OCSORT}, StrongSORT~\cite{strongsort}, BoTSORT~\cite{botsort}, and ByteTrack~\cite{bytetrack}. All trackers used the same YOLOv8 detector.
	
	\subsubsection{Qualitative Analysis}
\begin{figure*}[htp]
		\centering
		\includesvg[width=0.95\textwidth, inkscapearea=drawing]{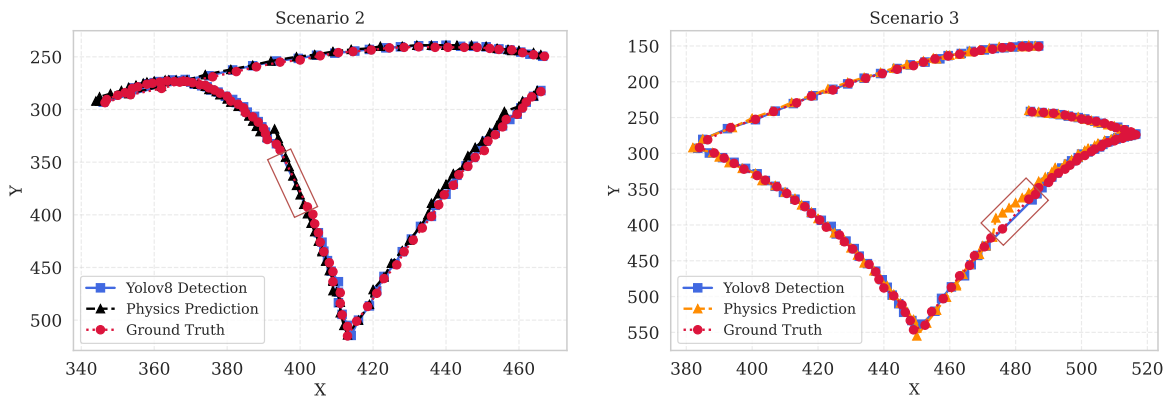}
		\vspace{-0.5cm}
		\caption{\centering Comparison of trajectories between object detection method, ground truth and our physics model. In both of scenarios our model can interpolate these gaps, using kinematics motion model, to produce continuous trajectory. The tiny rectangular boxes show interpolated positions.}
		\label{missedDet}
	\end{figure*}

	Figure~\ref{missedDet} illustrates how our physics-based tracking model constructs nearly perfect trajectories by recovering missing data points. The smooth trajectory demonstrates the model's ability to maintain tracking continuity even during detection failures. 
	
\begin{figure*}[ht]
		\centering
		\includesvg[width=0.90\textwidth]{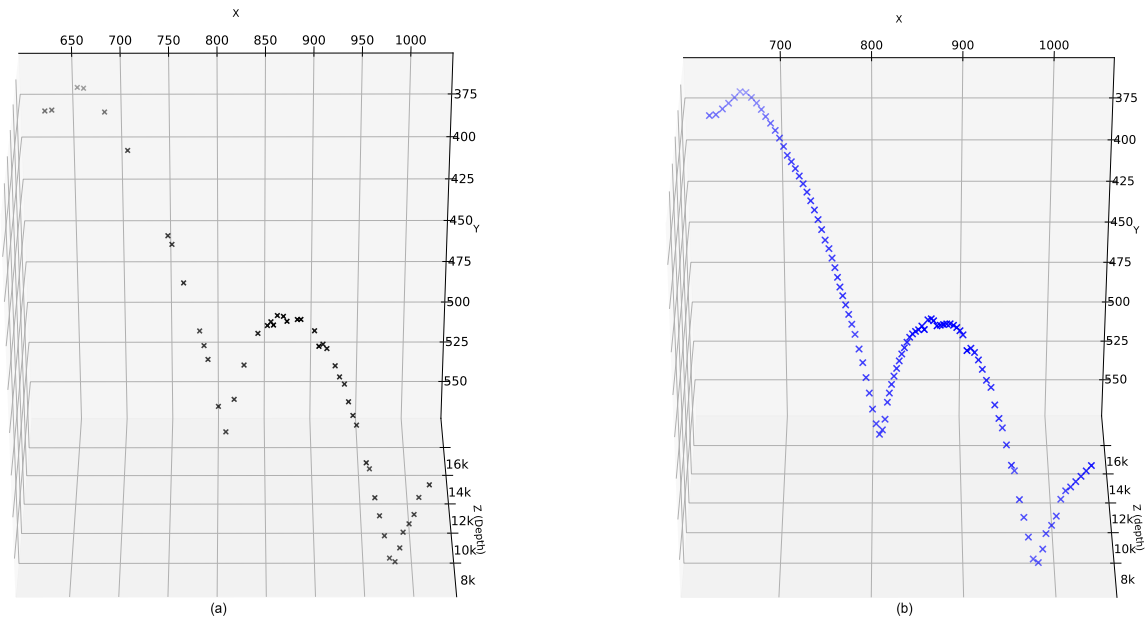}
		\vspace{-2cm}
		\caption{\centering Plot (a) shows a fragmented 3D trajectory of the ball in scenario 4. The fragmented trajectory is a result of missed detections due to occlusion and high speed. Plot (b) shows the reconstructed trajectory of the ball. We use a kinematics motion model forecast the position of the ball, when object detection fails, leading to smooth accurate trajectory.}
		\label{fig:sidebyside}
	\end{figure*}

	Figure~\ref{fig:sidebyside} provides a comparison between incomplete trajectories with missing points plot \textbf{a}, and the complete trajectory reconstructed by our physics-based model plot \textbf{b}. The 3D visualization highlights how our system is able to maintain accurate spatial positioning even during challenging scenarios. Our system's ability to forecast the trajectory plays crucial role in filling fragmented trajectories.
	
	\subsubsection{Quantitative Comparison}
	Table~\ref{OverallTracking} presents a comprehensive comparison between deep learning-based trackers and our physics-based approach across five challenging scenarios - Scenario 1: Multiple rapid bounces, Scenario 2: Wall collision at high speed, Scenario 3: Partial occlusion, Scenario 4: Distance variation (far to near), and Scenario 5: mixed collision with floor and side walls. 
	
	Our physics-based tracker consistently outperformed all competitors across all scenarios, achieving significant reductions in tracking error:
	\begin{itemize}[itemsep=0.2em]
		\item 63-73\% lower ADE compared to DeepOCSORT (best KF-based tracker)
		\item 68-90\% lower ADE compared to StrongSORT (worst performer)
		\item Similar patterns for AMD, with our approach achieving 65-80\% error reduction
	\end{itemize}
	
	Notably, ByteTrack and BoTSORT, despite strong performance on standard benchmarks, failed completely in some scenarios (marked by empty cells), particularly when the ball hit sidewalls or underwent multiple bounces. This failure highlights the inadequacy of standard motion models when dealing with the abrupt direction changes characteristic of racquetball trajectories. 
	
	DeepOCSORT performed best among the DL-based trackers but still showed substantial tracking drift during collision events. OCSORT and StrongSORT demonstrated reasonable tracking continuity but with significant positional errors, especially after collisions.
	
	\subsubsection{Scenario-Specific Analysis}
	Performance evaluation for each scenario reveals different thing like:
	\begin{itemize}[itemsep=0.2em]
		\item \textbf{Multiple Bounces (Scenario 1)}: Our approach achieved an ADE of 8.42 pixels compared to 22.97 for DeepOCSORT, demonstrating superior handling of sequential direction changes.
		\item \textbf{Wall Collisions (Scenario 2)}: All DL-based trackers struggled significantly with wall collisions, with errors 2-6× higher than our physics-based approach.
		\item \textbf{Occlusion Handling (Scenario 3)}: During partial occlusions, our physics model maintained tracking with an ADE of 48.32, slightly lower than DeepOCSORT's 50.0, using no appearance features.
		\item \textbf{Distance Variation (Scenario 4)}: When tracking objects moving between far and near positions, our approach maintained consistent accuracy (ADE 11.80) compared to highly variable performance from DL-based methods.
		\item \textbf{Mixed Collision (Scenario 5)}: In situations where there is mixed multiple collision with the floor and the side wall, our approach achieved remarkably low error (ADE 3.35), demonstrating robustness to detection uncertainty by leveraging physics-based motion constraints.
	\end{itemize}

	\definecolor{NavyBlue}{rgb}{0,0,0.545}
	\begin{table*}[ht]
		\centering
		\resizebox{0.95\linewidth}{!}{%
			\begin{tabular}{l*{10}{c}}
				\toprule
				& \multicolumn{2}{c}{\textbf{Scenario 1}} & \multicolumn{2}{c}{\textbf{Scenario 2}} & \multicolumn{2}{c}{\textbf{Scenario 3}} & \multicolumn{2}{c}{\textbf{Scenario 4}} & \multicolumn{2}{c}{\textbf{Scenario 5}} \\
				\cmidrule(lr){2-3} \cmidrule(lr){4-5} \cmidrule(lr){6-7} \cmidrule(lr){8-9} \cmidrule(lr){10-11}
				\textbf{Method} & ADE & AMD & ADE & AMD & ADE & AMD & ADE & AMD & ADE & AMD \\
				\midrule
				DeepOCSORT      & 22.97 & 0.69 & 31.31 & 0.91 & 50.42 & 0.91 & 19.90 & 0.87 & 10.80 & 0.15 \\
				OCSORT          & 65.26 & 1.83 & 93.50 & 2.54 & 111.80 & 2.54 & 39.16 & 2.09 & 89.87 & 1.44 \\
				StrongSORT      & 99.82 & 2.86 & 169.70 & 23.90 & 157.22 & 23.90 & 153.70 & 2.84 & 104.30 & 1.56 \\
				BOTSORT         & 51.70 & 1.81 & 158.00 & 11.59 & 74.11 & 11.59 & 178.50 & 1.67 & 70.20 & 0.72 \\
				Bytetrack       & 39.40 & 3.22 & 146.40 & 13.02 & 56.48 & 13.02 & 215.00 & 2.39 & -- & -- \\
				\midrule
				\textbf{Physics-based (Ours)} & \textbf{8.42} & \textbf{0.22} & \textbf{26.70} & \textbf{0.81} & \textbf{48.32} & \textbf{0.86} & \textbf{11.80} & \textbf{0.50} & \textbf{3.35} & \textbf{0.044} \\
				\bottomrule
			\end{tabular}
		}
		\caption{Performance comparison between DL-based and Physics-based trackers. Our model achieves best performance across all scenarios.}
		\label{OverallTracking}
	\end{table*}
	
	These results confirm that explicit modeling of physical motion, particularly collision dynamics, provides substantial advantages for tracking fast-moving objects with erratic trajectories. While Kalman filter-based approaches assume relatively linear motion between observations, our physics-based model accurately predicts the non-linear trajectories resulting from collisions, enabling more robust tracking even during temporary detection failures.
	
	\section{Conclusion}
	
	This research addresses the significant challenge of detecting and tracking fast-moving small objects, a problem that has received limited attention despite its importance in many applications. We proposed a comprehensive system combining deep learning-based detection with physics-based tracking to overcome the limitations of existing approaches when applied to objects like racquetballs, which move at high speeds and undergo frequent direction changes. 
	
	Our experiment shows that while recent advances in object detection have improved general detection performance, small fast-moving objects remain challenging. Among tested detectors, YOLOv8 gives the best balance between accuracy and real-time performance, achieving sufficient detection accuracy while maintaining 41 FPS even at high resolution. 
	
	The most significant contribution of this work is the physics-based tracking approach, which substantially outperforms state-of-the-art deep learning-based trackers across all test scenarios. By incorporating kinematic motion equations and collision dynamics, our tracker achieves 63-90\% lower tracking error compared to the best deep learning alternatives. This improvement is particularly pronounced during collision events, where traditional trackers struggle with the abrupt direction changes. 
	
	The integration of depth information further enhances tracking robustness, enabling true 3D trajectory modeling and providing an additional dimension for outlier detection and validation. This 3D awareness represents a significant advancement over existing approaches that operate solely in 2D space. 
	
	Our system design and methodology provide a foundation for improved perception capabilities in applications requiring accurate tracking of small, fast-moving objects, such as robotics, sports analysis, and autonomous systems.
	
	\section{Future Works and Limitations}

	While our approach demonstrates significant improvements over existing methods, several limitations and opportunities for future work remain in detection enhancements, advanced physics modeling, dataset expansion, and robotics integration among many.
	
	\subsection{Detection Enhancement}
	Future work could explore hybrid detection approaches combining deep learning with traditional computer vision techniques. Our preliminary experiments with Hough Transform-based detection showed promise in specific scenarios but lacked the generalizability of deep learning approaches. A combined approach could potentially leverage the strengths of both paradigms -- deep learning method for robust detection and hough transform or other geometric methods for scenarios where the object's shape is distinctive. We could also benefit from temporal integration to leverage motion information during the detection phase.
	
	\subsection{Advanced Physics Modeling}
	Our current kinematics-based tracker operates with simplified assumptions about external forces. A more comprehensive dynamics-based model could incorporate additional factors like -- air resistance and drag effects, spin-induced forces (magnus effect), surface interaction variation based on material properties, and energy dissipation models for more accurate prediction of bounce. \par
	Such enhancements would further improve trajectory prediction accuracy, particularly over extended sequences with multiple collisions.
	
	\subsection{Dataset Expansion}
	The current dataset, while specialized for our application, has limitations in scale and scenario diversity. We can expand the dataset to include greater variety of lighting conditions, multiple object tracking scenarios, different court materials and configurations and various ball types with different physical properties. 
	
	
	\subsection{Robotics Integration}
	Implementing our tracking system on robotic platforms represents a natural extension of this work. Integration with platforms like omnidirectional robots or turtle bots would provide real-world validation of the system's capabilities and enable research into closed-loop perception-action systems for interacting with fast-moving objects. This integration would require addressing additional challenges like, real-time processing constraints on embedded hardware, inferencing latency management for high-speed interactions, and motion planning and control strategies for intercepting predicted trajectories. \par
	
	Addressing these limitations and pursuing these future directions would further advance the state of the art in fast-moving small object detection and tracking, enabling new applications across robotics, sports analytics, and autonomous systems.
	
	{\small
		\bibliographystyle{ieee_fullname}
		\bibliography{egbib}

\begin{thebibliography}{10}\itemsep=-1pt

\bibitem{botsort}
Nir Aharon, Roy Orfaig, and Ben-Zion Bobrovsky.
\newblock Bot-sort: Robust associations multi-pedestrian tracking.
\newblock {\em arXiv preprint arXiv:2206.14651}, 2022.

\bibitem{yolo-z}
Aduen Benjumea, Izzeddin Teeti, Fabio Cuzzolin, and Andrew Bradley.
\newblock Yolo-z: Improving small object detection in yolov5 for autonomous
  vehicles.
\newblock {\em arXiv preprint arXiv:2112.11798}, 2021.

\bibitem{SORT}
Alex Bewley, Zongyuan Ge, Lionel Ott, Fabio Ramos, and Ben Upcroft.
\newblock Simple online and realtime tracking.
\newblock In {\em 2016 IEEE international conference on image processing
  (ICIP)}, pages 3464--3468. IEEE, 2016.

\bibitem{OCSORT}
Jinkun Cao, Jiangmiao Pang, Xinshuo Weng, Rawal Khirodkar, and Kris Kitani.
\newblock Observation-centric sort: Rethinking sort for robust multi-object
  tracking.
\newblock In {\em Proceedings of the IEEE/CVF Conference on Computer Vision and
  Pattern Recognition}, pages 9686--9696, 2023.

\bibitem{surveySOD}
Gong Cheng, Xiang Yuan, Xiwen Yao, Kebing Yan, Qinghua Zeng, Xingxing Xie, and
  Junwei Han.
\newblock Towards large-scale small object detection: Survey and benchmarks.
\newblock {\em IEEE Transactions on Pattern Analysis and Machine Intelligence},
  2023.

\bibitem{meanshift1}
Dorin Comaniciu, Visvanathan Ramesh, and Peter Meer.
\newblock Real-time tracking of non-rigid objects using mean shift.
\newblock In {\em Proceedings IEEE Conference on Computer Vision and Pattern
  Recognition. CVPR 2000 (Cat. No. PR00662)}, volume~2, pages 142--149. IEEE,
  2000.

\bibitem{DOTA}
Jian Ding, Nan Xue, Gui-Song Xia, Xiang Bai, Wen Yang, Michael~Ying Yang, Serge
  Belongie, Jiebo Luo, Mihai Datcu, Marcello Pelillo, et~al.
\newblock Object detection in aerial images: A large-scale benchmark and
  challenges.
\newblock {\em IEEE transactions on pattern analysis and machine intelligence},
  44(11):7778--7796, 2021.

\bibitem{strongsort}
Yunhao Du, Zhicheng Zhao, Yang Song, Yanyun Zhao, Fei Su, Tao Gong, and
  Hongying Meng.
\newblock Strongsort: Make deepsort great again.
\newblock {\em IEEE Transactions on Multimedia}, 2023.

\bibitem{everett2004dynamics}
Louis~J Everett and Alan~A Barhorst.
\newblock {\em Dynamics for Engineering Practice}.
\newblock McGraw-Hill Primis Custom Publishing, 2004.

\bibitem{sift4}
Saeid Fazli, Hamed~Moradi Pour, and Hamed Bouzari.
\newblock Particle filter based object tracking with sift and color feature.
\newblock In {\em 2009 Second international conference on machine vision},
  pages 89--93. IEEE, 2009.

\bibitem{fastrcnn}
Ross Girshick.
\newblock Fast r-cnn, 2015.

\bibitem{tracknet}
Yu-Chuan Huang, I-No Liao, Ching-Hsuan Chen, Ts{\`\i}-U{\'\i} {\.I}k, and
  Wen-Chih Peng.
\newblock Tracknet: A deep learning network for tracking high-speed and tiny
  objects in sports applications.
\newblock In {\em 2019 16th IEEE International Conference on Advanced Video and
  Signal Based Surveillance (AVSS)}, pages 1--8. IEEE, 2019.

\bibitem{sift3}
Falah Jabar, Sajad Farokhi, and UU Sheikh.
\newblock Object tracking using sift and klt tracker for uav-based
  applications.
\newblock In {\em 2015 IEEE International Symposium on Robotics and Intelligent
  Sensors (iris)}, pages 65--68. IEEE, 2015.

\bibitem{ju2020trajectory}
Nyan-Ping Ju, Dung-Ru Yu, Tsi-Ui Ik, and Wen-Chih Peng.
\newblock Trajectory-based badminton shots detection.
\newblock In {\em 2020 International Conference on Pervasive Artificial
  Intelligence (ICPAI)}, pages 64--71. IEEE, 2020.

\bibitem{khanam2024yolov5deeplookinternal}
Rahima Khanam and Muhammad Hussain.
\newblock What is yolov5: A deep look into the internal features of the popular
  object detector, 2024.

\bibitem{kisantal2019augmentation}
Mate Kisantal.
\newblock Augmentation for small object detection.
\newblock {\em arXiv preprint arXiv:1902.07296}, 2019.

\bibitem{kuhn1955hungarian}
Harold~W Kuhn.
\newblock The hungarian method for the assignment problem.
\newblock {\em Naval research logistics quarterly}, 2(1-2):83--97, 1955.

\bibitem{DeepOC}
Gerard Maggiolino, Adnan Ahmad, Jinkun Cao, and Kris Kitani.
\newblock Deep oc-sort: Multi-pedestrian tracking by adaptive
  re-identification.
\newblock {\em arXiv preprint arXiv:2302.11813}, 2023.

\bibitem{marvasti2020comet}
Seyed~Mojtaba Marvasti-Zadeh, Javad Khaghani, Hossein Ghanei-Yakhdan, Shohreh
  Kasaei, and Li Cheng.
\newblock Comet: Context-aware iou-guided network for small object tracking.
\newblock In {\em Proceedings of the Asian Conference on Computer Vision},
  2020.

\bibitem{meanshift3}
Jifeng Ning, Lei Zhang, David Zhang, and Chengke Wu.
\newblock Scale and orientation adaptive mean shift tracking.
\newblock {\em IET Computer Vision}, 6(1):52--61, 2012.

\bibitem{Metrics}
Rafael Padilla, Wesley Lobato~Passos, Thadeu Dias, Sergio Netto, and Eduardo da
  Silva.
\newblock A comparative analysis of object detection metrics with a companion
  open-source toolkit.
\newblock {\em Electronics}, 10:279--306, 01 2021.

\bibitem{fasterrcnn}
Shaoqing Ren, Kaiming He, Ross Girshick, and Jian Sun.
\newblock Faster r-cnn: Towards real-time object detection with region proposal
  networks.
\newblock {\em IEEE transactions on pattern analysis and machine intelligence},
  39(6):1137--1149, 2016.

\bibitem{psingh2025}
Prithvi~Raj Singh, Raju Gottumukkala, and Anthony Maida.
\newblock An analysis of kalman filter based object tracking methods for
  fast-moving tiny objects, 2025.

\bibitem{sun2020tracknetv2}
Nien-En Sun, Yu-Ching Lin, Shao-Ping Chuang, Tzu-Han Hsu, Dung-Ru Yu, Ho-Yi
  Chung, and Ts{\`\i}-U{\'\i} {\.I}k.
\newblock Tracknetv2: Efficient shuttlecock tracking network.
\newblock In {\em 2020 International Conference on Pervasive Artificial
  Intelligence (ICPAI)}, pages 86--91. IEEE, 2020.

\bibitem{meanshift2}
Tomas Vojir, Jana Noskova, and Jiri Matas.
\newblock Robust scale-adaptive mean-shift for tracking.
\newblock {\em Pattern Recognition Letters}, 49:250--258, 2014.

\bibitem{wang2023yolov7}
Chien-Yao Wang, Alexey Bochkovskiy, and Hong-Yuan~Mark Liao.
\newblock Yolov7: Trainable bag-of-freebies sets new state-of-the-art for
  real-time object detectors.
\newblock In {\em Proceedings of the IEEE/CVF conference on computer vision and
  pattern recognition}, pages 7464--7475, 2023.

\bibitem{deepSORT}
Nicolai Wojke, Alex Bewley, and Dietrich Paulus.
\newblock Simple online and realtime tracking with a deep association metric,
  2017.

\bibitem{yaseen2024yolov8indepthexplorationinternal}
Muhammad Yaseen.
\newblock What is yolov8: An in-depth exploration of the internal features of
  the next-generation object detector, 2024.

\bibitem{tinyperson}
Xuehui Yu, Yuqi Gong, Nan Jiang, Qixiang Ye, and Zhenjun Han.
\newblock Scale match for tiny person detection.
\newblock In {\em Proceedings of the IEEE/CVF winter conference on applications
  of computer vision}, pages 1257--1265, 2020.

\bibitem{golfball}
Xiaohan Zhang, Tianxiao Zhang, Yiju Yang, Zongbo Wang, and Guanghui Wang.
\newblock Real-time golf ball detection and tracking based on convolutional
  neural networks.
\newblock In {\em 2020 IEEE International Conference on Systems, Man, and
  Cybernetics (SMC)}, pages 2808--2813. IEEE, 2020.

\bibitem{bytetrack}
Yifu Zhang, Peize Sun, Yi Jiang, Dongdong Yu, Fucheng Weng, Zehuan Yuan, Ping
  Luo, Wenyu Liu, and Xinggang Wang.
\newblock Bytetrack: Multi-object tracking by associating every detection box.
\newblock In {\em European Conference on Computer Vision}, pages 1--21.
  Springer, 2022.

\bibitem{sift2}
Gangqiang Zhao, Ling Chen, Jie Song, and Gencai Chen.
\newblock Large head movement tracking using sift-based registration.
\newblock In {\em Proceedings of the 15th ACM international conference on
  Multimedia}, pages 807--810, 2007.

\bibitem{SIFT1}
Huiyu Zhou, Yuan Yuan, and Chunmei Shi.
\newblock Object tracking using sift features and mean shift.
\newblock {\em Computer vision and image understanding}, 113(3):345--352, 2009.

\bibitem{visDrone}
Pengfei Zhu, Longyin Wen, Dawei Du, Xiao Bian, Heng Fan, Qinghua Hu, and Haibin
  Ling.
\newblock Detection and tracking meet drones challenge.
\newblock {\em IEEE Transactions on Pattern Analysis and Machine Intelligence},
  44(11):7380--7399, 2021.

\end{thebibliography}
	}
	
\end{document}